%% file: main.tex
\pdfoutput=1

\documentclass[11pt]{article}

\usepackage{EMNLP2022}

\usepackage[T1]{fontenc}

\usepackage[utf8]{inputenc}

\usepackage{microtype}

\usepackage{inconsolata}

\usepackage{times}
\usepackage{latexsym}
\usepackage{amsmath,amssymb}
\usepackage{cleveref}
\usepackage{multirow}
\usepackage{graphicx}
\usepackage{caption}
\usepackage{subcaption}
\usepackage{booktabs}
\usepackage{url}
\usepackage{paralist}
\usepackage{tablefootnote}
\usepackage{color}
\usepackage{makecell}
\usepackage{enumitem}
\usepackage[normalem]{ulem}
\usepackage{stfloats}
\usepackage{colortbl}

\newcommand{\mbf}[1]{\boldsymbol{\mathbf{#1}}}
\newcommand{\mscript}[1]{\text{\scriptsize{#1}}}
\crefname{equation}{Eq.}{Eq.}
\crefname{section}{Sec.}{Sec.}

\title{
    Mitigating Data Sparsity for Short Text Topic Modeling \\ by Topic-Semantic Contrastive Learning
}

\author{
  Xiaobao Wu$^\ddagger$ \quad Anh Tuan Luu$^\ddagger$ \quad Xinshuai Dong$^\dagger$  \\
  $^\ddagger$Nanyang Technological University \quad
  $^\dagger$Carnegie Mellon University \\
  \texttt{xiaobao002@e.ntu.edu.sg}, \quad
  \texttt{anhtuan.luu@ntu.edu.sg} \\
  \texttt{xinshuad@andrew.cmu.edu}
}

\begin{document}
    \maketitle
    \begin{abstract}
        To overcome the data sparsity issue in short text topic modeling,
        existing methods commonly rely on data augmentation or the data characteristic of short texts
        to introduce more word co-occurrence information.
        However, most of them do not make full use of the augmented data or the data characteristic:
        they insufficiently learn the relations among samples in data,
        leading to dissimilar topic distributions of semantically similar text pairs.
        To better address data sparsity, in this paper
        we propose a novel short text topic modeling framework, Topic-Semantic Contrastive Topic Model (TSCTM).
        To sufficiently model the relations among samples,
        we employ a new contrastive learning method with efficient positive and negative sampling strategies based on topic semantics.
        This contrastive learning method refines the representations, enriches the learning signals, and thus mitigates the sparsity issue.
        Extensive experimental results show that our TSCTM outperforms state-of-the-art baselines
         regardless of the data augmentation availability, producing high-quality topics and topic distributions.
        \footnote{Our code is available at \url{https://github.com/bobxwu/TSCTM}.}
    \end{abstract}

    \section{Introduction}
        Topic models aim to discover the latent topics of a document collection and infer the topic distribution of each document 
        in an unsupervised fashion
        \cite{blei2003latent}.
        Due to the effectiveness and interpretability,
        topic models have been popular for decades with various downstream applications \cite{Ma2012,mehrotra2013improving,boyd2017applications}.
        However, despite the success on long texts,
        current topic models generally cannot handle well short texts, such as tweets, headlines, and comments \cite{yan2013biterm}.
        The reason lies in that topic models
        rely on word co-occurrence information to infer latent topics,
        but such information is extremely scarce in short texts  \cite{qiang2020short}.
        This issue, referred to as \textit{data sparsity},
        can hinder state-of-the-art topic models from discovering high-quality topics
        and thus has attracted much attention.

\input{input/fig_motivation}

        To overcome the data sparsity issue,
        traditional wisdom can be mainly categorized into two lines:
        \begin{inparaenum}[(i)]
            \item
                Augment datasets with more short texts containing similar semantics \cite{phan2008learning,jin2011transferring,chen2015word}.
                This way can feed extra word co-occurrence information to topic models.
            \item
                Due to the limited context, many short texts in the same collection, such as tweets from Twitter, tend to be relevant, sharing similar topic semantics \cite{qiang2020short};
                to leverage this data characteristic,  models such as
                 DMM \cite{yin2014dirichlet,Li2016} and state-of-the-art NQTM \cite{Wu2020short}
                 learn similar topic distributions from relevant samples.
        \end{inparaenum}
        These two lines of thought have been shown to achieve good performance and mitigate data sparsity to some extent.

        However,  existing short text topic models neither make full use of the augmented data nor the crucial data characteristic.
        To begin with, an augmented text is expected to have a similar topic distribution as the original text since they share similar topic semantics,
        but existing approaches tend to overlook this important relation between samples.
        As shown in \Cref{fig_motivation_baseline},
        text $\mbf{x}^{(i)}$ and its augmented view $\mbf{x}^{(i)}_{+}$ have similar topic semantics,
        but their topic distributions inferred by NQTM are far~from~similar.
        Moreover, guided by the aforementioned data characteristic,
            state-of-the-art methods like NQTM
            attempt to learn similar topic distributions for relevant samples,
            yet they could inappropriately do so.
        \Cref{fig_motivation_baseline} shows that
        text $\mbf{x}^{(i)}$ and $\mbf{x}^{(\ell)}$ are relevant,
        but their learned topic distributions are dissimilar;
        $\mbf{x}^{(i)}$ and $\mbf{x}^{(j)}$ are irrelevant,
        but theirs are similar.
        In a word,
        current approaches insufficiently model the relations among samples in data,
        which hinders fully addressing the data sparsity issue.

        To better mitigate data sparsity,
        we in this paper propose Topic-Semantic Contrastive Topic Model (TSCTM),
        a novel short text topic modeling framework that unifies both cases with and without data augmentation.
        To be specific,
        TSCTM makes full use of relations among samples with a novel topic-semantic contrastive learning method.
        In the case without data augmentation,
        TSCTM effectively samples positive and negative text pairs based on topic semantics.
        In the case with data augmentation,
        TSCTM also smoothly incorporates the relations between augmented and original samples, enabling better utilization of data augmentation.
        Through the novel contrastive learning method,
        TSCTM sufficiently models the relations among samples,
        which enriches the learning signals, refines the learning of representations,
        and thus mitigates the data sparsity issue
        (see \Cref{fig_motivation_our} for an illustration).
        We summarized the main contributions of this paper as follows:
        \begin{itemize}
            \item
                We follow a contrastive learning perspective
                and propose a novel contrastive learning method with efficient positive and negative pairs sampling strategies
                to address the data sparsity issue in short text topic modeling.
            \item
                We propose a novel short text topic modeling framework,
                Topic-Semantic Contrastive Topic Model (TSCTM),
                which is the first such framework that concerns both cases with and without data augmentation.
            \item
                We validate our method with extensive experiments where TSCTM effectively mitigates data sparsity 
                and consistently surpasses state-of-the-art baselines, producing high-quality topics and topic distributions.
        \end{itemize}

\input{input/sec_related_work}

    \section{Methodology}
        In this section, we first review the background of topic modeling.
        Then we introduce topic-semantic contrastive learning, a novel approach for short text topic modeling.
        Finally, we put this contrastive learning into the topic modeling context and propose our Topic-Semantic Contrastive Topic Model.

        \subsection{Notations and Problem Setting}
            Our notations and problem setting of topic modeling follow LDA \cite{blei2003latent}.
            Consider a collection of $N$ documents $\{\mbf{x}^{(1)}, \dots, \mbf{x}^{(N)}\}$ with $V$ unique words, \emph{i.e.}, vocabulary size.
            We require to discover $K$ topics from the collection.
            Each topic is interpreted as its relevant words and defined as a distribution over all words (topic-word distribution): $\mbf{\beta}_{k} \in \mathbb{R}^{V} $.
            Then, $\mbf{\beta} \! = \! (\mbf{\beta}_{1}, \dots, \mbf{\beta}_{K}) \! \in \! \mathbb{R}^{V \times K} $ is the topic-word distribution matrix.
            A topic model also infers what topics a document contains, \emph{i.e.}, the topic distribution of a document, denoted as $\mbf{\theta} \in \Delta_{K}$.~\footnote{Here $\Delta_{K}$ denotes a probability simplex defined as  $\Delta_{K} = \{ \mbf{\theta} \in \mathbb{R}_{+}^{K} | \sum_{k=1}^{K}\theta_{k} = 1 \}$.}

        \subsection{Topic-Semantic Contrastive Learning}
            The core difference between our TSCTM and a conventional topic model lies in that
            we employ the novel topic-semantic contrastive learning method
            to model the relations among samples.
            As such, the learning signals are enriched through sufficiently modeling the relations among texts to address the data sparsity issue.
            \Cref{fig_illustration} illustrates our topic-semantic contrastive learning method.

            \subsubsection{Encoding Short Texts}
                To employ our topic-semantic contrastive learning,
                the first step is to encode short text inputs into a semantic space and obtain the corresponding representations and topic distributions.
                Specifically,
                we employ an encoder neural network $f_{\Theta}$ with parameter $\Theta$
                to encode short text $\mbf{x}^{(i)}$ and get its representation $\mbf{h}^{(i)} = f_{\Theta} (\mbf{x}^{(i)})$.
                The topic distribution of $\mbf{x}^{(i)}$ is denoted as  $\mbf{\theta}^{(i)}$ 
                and is computed by normalizing $\mbf{h}^{(i)}$ into a probability simplex with a softmax function
                as $\mbf{\theta}^{(i)} = \mathrm{softmax}(\mbf{h}^{(i)}) $.
                Note that we train topic distribution $\mbf{\theta}^{(i)}$ with a topic modeling objective, which will be introduced later.

                \input{input/fig_illustration}

            \subsubsection{Positive Pairs for Contrastive Learning} \label{sec_modeling_positive}
                To utilize the vital characteristic of short texts
                (many short texts in a collection like Twitter tend to share similar topics due to the limited context), 
                we propose to find those semantically similar texts and model them as positive pairs to each other for contrastive learning.
                Therefore, we can employ a contrastive learning objective to align those semantically similar texts in terms of representations and thus topic distributions.

                However, it is non-trivial to find those semantically similar texts as positive pairs.
                Some previous methods like 
                CLNTM \cite{nguyen2021contrastive} samples salient words to build positive pairs for long texts,
                but this way does not fit short texts well due to the extremely limited context (shown in \Cref{sec_topic_quality}).
                Differently,
                DMM \cite{yin2014dirichlet,Li2016} follows a clustering process to aggregate short texts with similar topics,
                but lacks the flexibility of model design as it requires model-specific derivations for parameter inference.
                As such, we propose to employ vector quantization \cite{VandenOord2017} to find positive pairs for short texts.

                Specifically, 
                as shown in \Cref{fig_illustration},
                we first quantize topic distribution $\mbf{\theta}^{(i)}$ to the closest embedding vector,
                and its quantized  topic distribution $\mbf{\theta}_{q}^{(i)}$ is computed as:
                \begin{align}
                    \mbf{\theta}_{q}^{(i)} &= \mbf{e}_{q(\mbf{\theta}^{(i)})} \\
                    q(\mbf{\theta}^{(i)}) &= \mathrm{argmin}_{k} \| \mbf{\theta}^{(i)} - \mbf{e}_{k} \|_{2}. \label{eq_quantization}
                \end{align}
                Here, $( \mbf{e}_{1}, \mbf{e}_{2}, \dots, \mbf{e}_{K} ) \in \mathbb{R}^{K \times K}$ are $K$ predefined embedding vectors,
                 and $q(\cdot) \! \in \! \{1, \dots, K\} $ outputs the index of the quantized embedding vector.
                These embedding vectors are initialized as different one-hot vectors before training to ensure that they are far away from each other for distinguishable quantization \cite{Wu2020short}.
                We then model the short texts with the same quantization indices as positive pairs, as follows:
                \begin{equation}
                    \{ \mbf{x}^{(i)}, \mbf{x}^{(\ell)} \} \quad \text{where} \quad q(\mbf{\theta}^{(\ell)}) = q(\mbf{\theta}^{(i)}).
                    \label{eq_positive_pair}
                \end{equation}
                This is because topic distributions of short texts with similar semantics are learned to be quantized to the same embedding vectors.

            \subsubsection{Negative Pairs for Contrastive Learning} \label{sec_modeling_negative}
                We first explain why we need to push negative pairs away from each other.
                Then we propose a novel semantic-based negative sampling strategy to sample semantically effective negative pairs.

                \paragraph{Why Negative Pairs?} \label{sec_problem_positive}
                We also need negative pairs to sufficiently model the relations among samples.
                Pulling close semantically similar short texts provides additional learning signals to address data sparsity, 
                however two texts with different semantics can sometimes be wrongly viewed as a positive pair, leading to less distinguishable representations (see \Cref{fig_motivation_baseline}).
                To mitigate this issue, we propose to find negative pairs in the data and explicitly push them away, so we can sufficiently model the relations among samples to better improve topic modeling for short texts.
                    The use of negative pairs can also  be supported from an information-theoretical perspective following \citet{wang2020understanding}:
                    pushing away negative pairs facilitates uniformity, thus maximizing the mutual information of the representations of positive pairs.
                    Otherwise, if we only pull close positive pairs,
                    chances are high that all the representations will collapse towards each other and become less distinguishable.

                    In a word, pulling close positive pairs and pushing away negative pairs are both vital for better representations and topic distributions,
                    and they together justify the use of contrastive learning
                    to regularize the learning of short text topic models 
                    (see empirical support in \Cref{sec_topic_quality,sec_ablation_study}).

                \paragraph{Semantic-based Negative Sampling}
                    Conventional contrastive learning methods such as \citet{he2020momentum,chen2020simple}
                    simply take different samples as negative pairs.                %
                    This is reasonable in the context of long text topic modeling as different samples in a long text dataset have sufficiently various contexts to contain different topics.
                    However, for  short text topic modeling, 
                    many samples actually share similar topics as the aforementioned data characteristic.
                    Therefore,
                    simply taking different samples as negative pairs can wrongly push away semantically similar pairs,
                    which hampers topic modeling performance (shown in \Cref{sec_ablation_study}).

                    To overcome this issue, we here propose a neat and novel semantic-based negative sampling strategy.
                    Similar to our positive pair sampling strategy,
                    we sample negative pairs according to the quantization result as in \Cref{eq_quantization}.
                    Specifically,
                    two texts are expected to contain different topics semantics if their topic distributions are quantized to different embedding vectors;
                    thus we take such a pair of texts as a negative pair $\{\mbf{x}^{(i)}, \mbf{x}^{(j)} \}$:
                    \begin{equation}
                        \{\mbf{x}^{(i)}, \mbf{x}^{(j)} \} \quad \text{where} \quad q(\mbf{\theta}^{(j)}) \neq q(\mbf{\theta}^{(i)}).
                        \label{eq_negative_pair}
                    \end{equation}
                    Our negative sampling strategy better aligns with the characteristic of short texts,
                    and does not introduce complicated preprocessing steps or additional modules,
                    which simplifies the architecture and eases computational cost.

                \subsubsection{Topic-Semantic Contrastive Objective}
                    We have positive and negative pairs through our sampling strategies defined in \Cref{eq_positive_pair} and \Cref{eq_negative_pair}.
                    Now as illustrated in \Cref{fig_illustration}, we formulate our topic-semantic contrastive (TSC) objective following \citet{van2018representation}:
                    \begin{align}
                        & \mathcal{L}_{\mscript{TSC}} (\mbf{x}^{(i)}) = 
                        \sum_{\ell} \!\! -\log \! \frac{\exp{( g(\mbf{h}^{(i)}, \mbf{h}^{(\ell)})) }} {\sum_{j} \exp{( g( \mbf{h}^{(i)}, \mbf{h}^{(j)}) )}} ~, \notag \\
                        & \quad \quad \text{where} \; j \in \{ j' | q(\mbf{\theta}^{(j')}) \neq q(\mbf{\theta}^{(i)}) \} \notag \\
                        & \quad \quad \text{and} \; \ell \in \{ \ell' | q(\mbf{\theta}^{(\ell')}) = q(\mbf{\theta}^{(i)}) \} .
                        \label{eq_TSC}
                    \end{align}
                    In \Cref{eq_TSC}, $g(\cdot, \cdot)$ can be any score function to measure the  similarity between
                    two  representations, and we follow \citet{wu2018unsupervised} to employ the cosine similarity as $g(a, b) = \cos(a, b) / \tau $
                     where $\tau$ is a hyper-parameter controlling the scale of the score.
                    This objective pulls close the representations of positive pairs ($\mbf{h}^{(i)}, \mbf{h}^{(\ell)}$)
                     and pushes away the representations of negative pairs ($\mbf{h}^{(i)}, \mbf{h}^{(j)}$).
                    Thus this provides more learning signals to topic modeling by correctly capturing the
                     relations among samples, which alleviates the data sparsity issue.

            \subsection{Topic-Semantic Contrastive Topic Model}
                Now we are able to combine the topic-semantic contrastive objective with the objective of short text topic modeling
                to formulate our Topic-Semantic Contrastive Topic Model (TSCTM).

                \paragraph{Short Text Topic Modeling Objective}
                    We follow the framework of AutoEncoder to design our topic modeling objective.
                    As the input short text $\mbf{x}^{(i)}$ is routinely transformed into Bag-of-Words,
                    its reconstruction is modeled as sampling from a multinomial distribution: $\mathrm{Mult}(\mathrm{softmax}(\mbf{\beta}\mbf{\theta}_{q}^{(i)}))$ \cite{Miao2016}.
                    Here, $\mbf{\theta}_{q}^{(i)}$ is the quantized topic distribution for reconstruction,
                    and $\mbf{\beta}$ is a learnable parameter to model the topic-word distribution matrix.
                    Then, the expected log-likelihood is proportional to ${\mbf{x}^{(i)}}^{\top} \log ( \mathrm{softmax}(\mbf{\beta} \mbf{\theta}^{(i)}_{q}) ) $ \cite{Srivastava2017}.
                    Therefore, we define the objective for short text topic modeling (TM) as:
                     \begin{align}
                        &\mathcal{L}_{\text{TM}}(\mbf{x}^{(i)}) =  -{\mbf{x}^{(i)}}^{\top} \log ( \mathrm{softmax}(\mbf{\beta} \mbf{\theta}_{q}^{(i)}) ) \notag \\
                        & + \| \mathrm{sg}(\mbf{\theta}^{(i)}) - \mbf{\theta}_{q}^{(i)} \|_{2} + \lambda \| \mathrm{sg}(\mbf{\theta}_{q}^{(i)}) - \mbf{\theta}^{(i)} \|_{2}
                        \label{eq_TM}
                    \end{align}
                    where the first term measures the reconstruction error between the input and reconstructed text.
                    The last two terms refer to minimizing the distance between the topic distribution $\mbf{\theta}^{(i)}$ and quantized topic distribution $\mbf{\theta}_{q}^{(i)}$ respectively weighted by $\lambda$ \cite{VandenOord2017}.
                    Here $\mathrm{sg}(\cdot)$ denotes a stop gradient operation that prevents gradients from back-propagating to its inputs.
    
                \paragraph{Overall Learning Objective of TSCTM}
                    The overall learning objective of TSCTM is a combination of \Cref{eq_TM} and \Cref{eq_TSC}, as:
                    \begin{equation}
                         \mathcal{L}_{\mscript{TM}}(\mbf{x}^{(i)}) + \lambda_{\mscript{TSC}} \mathcal{L}_{\mscript{TSC}}(\mbf{x}^{(i)}),
                        \label{eq_overall}
                    \end{equation}
                    where $\lambda_{\mscript{TSC}}$ is a hyper-parameter controlling the weight of topic-semantic contrastive objective.
                    This learning objective can
                     learn meaningful representations from data and further refine the representations through modeling the relations among samples to enrich learning signals, which mitigates the data sparsity issue and improves the topic modeling performance of short texts.
    
            \subsection{Learning with Data Augmentation}
                In this section,
                 we adapt our Topic-Semantic Contrastive Topic Model to the case where data augmentation is available
                 to fully utilize the introduced augmentations.
                \paragraph{Incorporating Data Augmentation}
                    Let $\mbf{x}^{(i)}_{+}$ denote one augmented view of  $\mbf{x}^{(i)}$.
                    As our augmentation techniques can ensure that  $\mbf{x}^{(i)}$ and $\mbf{x}^{(i)}_{+}$ share similar topic semantics as much as possible
                    (details about how we augment data will be introduced in \Cref{sec_aug_tech}),
                    we explicitly consider $\mbf{x}^{(i)}$ and $\mbf{x}^{(i)}_{+}$ as a positive pair.
                    Besides, we consider $\mbf{x}^{(i)}$ and $\mbf{x}^{(j)}_{+}$ as a negative pair if $\mbf{x}^{(i)}$ and $\mbf{x}^{(j)}$ are so.
                    This is because if $\mbf{x}^{(i)}$ and $\mbf{x}^{(j)}$ possess dissimilar topic semantics,
                     then $\mbf{x}^{(i)}$ and $\mbf{x}^{(j)}_{+}$ should as well.
                    Taking these two points into consideration,
                        as shown in \Cref{fig_illustration},
                    we formulate our topic semantic contrastive objective with data augmentation as
                    \begin{align}
                        &\mathcal{L}_{\mscript{TSC}}(\mbf{x}^{(i)},\mbf{x}^{(i)}_{+}) = 
                        -\log \frac{ \exp{(g(\mbf{h}^{(i)}, \mbf{h}^{(i)}_{+}) )}}
                        {\mathcal{D}} \notag \\
                        & \quad \quad + \lambda_{\mscript{original}} \sum_{\ell} -\log \frac{\exp{(g( \mbf{h}^{(i)}, \mbf{h}^{(\ell)} ) )}}{ \mathcal{D} }~, \notag \\
                        & \mathcal{D} = \!\! {\sum_{j} \!\! \exp{(g(\mbf{h}^{(i)}, \mbf{h}^{(j)}) )} + \exp{(g(\mbf{h}^{(i)}, \mbf{h}^{(j)}_{+}) ) }  }~, \notag \\
                        & \quad \quad \text{where} \; j \in \{ j' | q(\mbf{\theta}^{(j')}) \neq q(\mbf{\theta}^{(i)}) \} \notag \\
                        & \quad \quad \text{and} \; \ell \in \{ \ell' | q(\mbf{\theta}^{(\ell')}) = q(\mbf{\theta}^{(i)}) \}.
                        \label{eq_TSC_DA}
                    \end{align}
                    Here $\lambda_{\mscript{original}}$ is a weight hyper-parameter of the contrastive objective for the positive pairs in the original dataset.
                    Compared to \Cref{eq_TSC},
                    this formulation additionally incorporates the relation between positive pair $\mbf{x}^{(i)}, \mbf{x}^{(i)}_{+}$ by making their representations $\mbf{h}^{(i)}$ and $\mbf{h}^{(i)}_{+}$ close to each other and the relation between negative pair $\mbf{x}^{(i)}, \mbf{x}^{(j)}_{+}$ by pushing away their representations $\mbf{h}^{(i)}$ and $\mbf{h}^{(j)}_{+}$.

    \input{input/tab_dataset}

                \paragraph{Overall Learning Objective of TSCTM with Data  Augmentation}
                    Combining \Cref{eq_TM} with augmented data and \Cref{eq_TSC_DA},
                    we are able to formulate the final learning objective of TSCTM with data augmentation as follows:
                    \begin{align}
                       \mathcal{L}_{\mscript{TM}}(\mbf{x}^{(i)}) + \mathcal{L}_{\mscript{TM}}(\mbf{x}^{(i)}_{+}) 
                         + \lambda_{\mscript{TSC}} \mathcal{L}_{\mscript{TSC}}(\mbf{x}^{(i)}, \mbf{x}^{(i)}_{+}),
                        \label{eq_overall_DA}
                    \end{align}
                    where we jointly reconstruct the positive pair $\mbf{x}^{(i)}, \mbf{x}^{(i)}_{+}$ and
                    regularize the learning by the topic-semantic contrastive objective with augmented samples.
                    Accordingly, our method smoothly adapts to the case with data augmentation.

\input{input/tab_topic_quality}

    \section{Experimental Setting}
        In this section, we conduct comprehensive experiments to show the effectiveness of our method.
        \subsection{Datasets}
            We employ the following benchmark short text datasets in our experiments:
            \begin{inparaenum}[(i)]
                \item
                    \textbf{TagMyNews title} contains news titles released by \citet{Vitale2012} with 7 annotated labels like ``sci-tech'' and ``entertainment''.
                \item
                    \textbf{AG News} includes news divided into 4 categories like ``sports'' and ``business'' \cite{Zhang2015}. We use the subset provided by \citet{rakib2020enhancement}.
                \item
                    \textbf{Google News} is from \citet{yin2014dirichlet} with 152 categories.
            \end{inparaenum}

            We preprocess datasets with the following steps \cite{Wu2020short}:
            \begin{inparaenum}[(i)]
                \item tokenize texts with nltk;~\footnote{\url{https://www.nltk.org/}}
                \item convert characters to lower cases;
                \item filter out illegal characters;
                \item remove texts with length less than 2;
                \item filter out low-frequency words.
            \end{inparaenum}
            The dataset statistics are reported in \Cref{tab_dataset}.

        \subsection{Data Augmentation Techniques} \label{sec_aug_tech}
            To generate augmented texts, we follow \citet{zhang2021supporting} and employ two simple and effective techniques:
            WordNet Augmenter and Contextual Augmenter.~\footnote{\url{https://github.com/makcedward/nlpaug}}
            WordNet Augmenter substitutes words in an input text with their synonymous selected from the WordNet database \cite{ma2019nlpaug}.
            Then, Contextual Augmenter leverages the pre-trained language models such as BERT \cite{devlin2018bert} to find the top-n suitable words of the input text for insertion or substitution \cite{kobayashi2018contextual,ma2019nlpaug}.
            To retain the original semantics as much as possible, we only change $30\%$ words and also filter low-frequency words following \citet{zhang2021supporting}.
            With these augmentation techniques, we can sufficiently retain original semantics and meanwhile bring in more word-occurrence information to alleviate the data sparsity of short texts.

        \subsection{Baseline Models}
            We compare our method with the following state-of-the-art baseline models:
            \begin{inparaenum}[(i)]
                \item \textbf{ProdLDA} \cite{Srivastava2017}~\footnote{\url{https://github.com/akashgit/autoencoding_vi_for_topic_models}}, a neural topic model based on the standard VAE with a logistic normal distribution as an approximation of Dirichlet prior.
                \item \textbf{WLDA} \cite{Nan2019}, a Wasserstein AutoEncoder \cite{tolstikhin2018wasserstein} based topic model.
                \item \textbf{CLNTM} \cite{nguyen2021contrastive}~\footnote{\url{https://github.com/nguyentthong/CLNTM}}, a recent topic model with contrastive learning designed for long texts, which samples salient words of texts as positive samples.
                \item \textbf{NQTM} \cite{Wu2020short}~\footnote{\url{https://github.com/bobxwu/NQTM}}, a state-of-the-art neural short text topic model with vector quantization.
                \item \textbf{WeTe} \cite{wang2022representing}~\footnote{\url{https://github.com/wds2014/WeTe}}, a recent state-of-the-art method using conditional transport distance to measure the reconstruction error between texts and topics which both are represented with embeddings.
            \end{inparaenum}
            Note that the differences between NQTM and our method are described in \Cref{sec_related_work}.
            The implementation detail of our method can be found in \Cref{sec_appendix_model_implementation}.

\input{input/tab_ablation_study}

    \section{Experimental Result}
        \subsection{Topic Quality Evaluation} \label{sec_topic_quality}
            \paragraph{Evaluation Metric}
            Following \citet{Nan2019,wang2022representing}, we evaluate the quality of discovered topics from two perspectives:
            \begin{inparaenum}[(i)]
                \item
                    \textbf{Topic Coherence}, meaning the words in a topic should be coherent.
                    We adopt the widely-used Coherence Value \cite[$C_V$, ][]{roder2015exploring} following \citet{Wu2020short}.
                    We use external Wikipedia documents \footnote{\url{https://github.com/dice-group/Palmetto}} as its reference corpus to estimate the co-occurrence probabilities of words.
                \item
                    \textbf{Topic Diversity}, meaning the topics should be distinct from each other instead of being repetitive. We use Topic Uniqueness \cite[$TU$, ][]{Nan2019} which measures the proportions of unique words in the discovered topics.
                    Hence a higher $TU$ score indicates the discovered topics are more diverse.
            \end{inparaenum}
            With these two metrics, we can comprehensively evaluate topic quality.
            We run each model 5 times and report the experimental results in the two cases: without and with data augmentation as follows.
            \paragraph{Without Data Augmentation}
                In the case without data augmentation, only original datasets are used for all models in the experiments, and our TSCTM uses \Cref{eq_overall} as the objective function.
                The results are reported in the upper part of \Cref{tab_topic_quality}.
                We see that TSCTM surpasses all baseline models in terms of both coherence ($C_V$) and diversity ($TU$) under 50 and 100 topics across all datasets.
                Besides, it is worth mentioning that our TSCTM significantly outperforms NQTM and CLNTM.
                NQTM insufficiently models the relations among samples since it only considers texts with similar semantics,
                and CLNTM samples salient words from texts for contrastive learning, which is ineffective for short texts with limited context.
                In contrast, our TSCTM can discover effective samples for learning contrastively based on the topic semantics, which sufficiently models the relations among samples, thus achieving higher performance.
                Note that examples of discovered topics are in \Cref{sec_appendix_topic_examples}.
                These results show that TSCTM is capable of producing higher-quality topics with better coherence and diversity.

        \paragraph{With Data Augmentation}
            In the case with data augmentation,
            we produce augmented texts to enrich datasets for all models through the techniques mentioned in \Cref{sec_aug_tech}, so all models are under the same data condition for fair comparisons.
            Note that our TSCTM uses \Cref{eq_overall_DA} as the objective function in this case.
            The results are summarized in the lower part of \Cref{tab_topic_quality}.
            We have the following observations:
            \begin{inparaenum}[(\bgroup \bfseries i\egroup)]
                \item
                    Data augmentation can mitigate the data sparsity issue of short text topic modeling to some extent.
                    \Cref{tab_topic_quality} shows that the topic quality of several baseline models is improved with augmentations compared to the case without.
                \item
                    TSCTM can better utilize augmentations and consistently achieves better topic quality performance.
                    As shown in \Cref{tab_topic_quality}, we see that TSCTM reaches the best $C_V$ and $TU$ scores compared to baseline models under 50 and 100 topics.
                    This shows that our method can better leverage augmentations through the new topic-semantic contrastive learning to further alleviate data sparsity and improve short text topic modeling.
            \end{inparaenum}

            The above results demonstrate that TSCTM can adapt to both cases with or without data augmentation,
            effectively overcoming the data sparsity challenge and producing higher-quality topics.

        \input{input/tab_clustering}

        \subsection{Ablation Study} \label{sec_ablation_study}
            We conduct an ablation study that manifests the effectiveness and necessity of our topic-semantic contrastive learning method.
            As shown in \Cref{tab_ablation_study},
            our TSCTM significantly outperforms the traditional contrastive learning \cite{chen2020simple} (w/ traditional contrastive).
            This shows the better effectiveness of our novel topic-semantic contrastive learning with the new positive and negative sampling strategies.
            Besides,
            if without modeling negative pairs (w/o negative pairs),
            the coherence ($C_V$) and diversity ($TU$) performance both greatly degrades, \emph{e.g.}, from 0.479 to 0.397 and from 0.969 to 0.503 on AG News.
            This is because only modeling positive pairs makes the representations all collapse together and become less distinguishable,
            which hinders the learning of topics and leads to repetitive and less coherent topics (see also \Cref{sec_modeling_negative}).
            Moreover,
            \Cref{tab_ablation_study} shows that
            the coherence performance is hampered
            in the case without positive pairs (w/o positive pairs).
            The reason lies in that the method cannot capture the relations between positive pairs to further refine representations,
            and thus the inferred topics become less coherent.
            These results show the effectiveness and necessity of the positive and negative sampling strategies of our topic-semantic contrastive learning method.

        \input{input/fig_classification}

        \subsection{Short Text Clustering}
            Apart from topic quality, we evaluate the quality of inferred topic distributions through short text clustering following \citet{wang2022representing}.
            Specifically, we use the most significant topics in the learned topic distributions of short texts as their cluster assignments.
            Then, we employ the commonly-used clustering metrics, \textbf{Purity} and
            \textbf{NMI}
            \cite{Manning2008} to measure the clustering performance as \citet{wang2022representing}.
            Note that our goal is not to achieve state-of-the-art clustering performance but to compare the quality of learned topic distributions.
            \Cref{tab_clustering} shows that the clustering performance of our model is generally the best over baseline models concerning both Purity and NMI.
            This demonstrates that our model can infer more accurate topic distributions of short texts.

        \subsection{Short Text Classification}
            In order to compare extrinsic performance, we conduct text classification experiments as a downstream task of topic models \cite{nguyen2021contrastive}.
            In detail, we use the learned topic distributions by different models as features and train SVM classifiers to predict the class of each short text.
            We use the labels from the adopted datasets.
            \Cref{fig_classification} shows that our TSCTM consistently achieves the best classification performance compared to baseline models.
            Note that the p-values of significance tests are all less than 0.05.
            This shows that the learned topic distributions of our model are more discriminative and accordingly can be better employed in the text classification downstream task.

        \input{input/fig_tsne}
        \input{input/tab_sim}

        \subsection{Analysis of Topic Distributions}
            In this section we analyze the learned topic distributions of short texts to evaluate the modeling of relations among samples.
            \Cref{fig_tsne} illustrates the t-SNE \cite{Maaten2008} visualization for the learned topic distributions of original and augmented short texts by ProdLDA, NQTM, and our TSCTM.
            It shows that the topic distributions learned by our TSCTM are more aggregated together and well separately scattered in the space, in terms of only original short texts or both original and augmented short texts.
            In addition, we report the cosine similarity between the topic distributions of original and augmented short texts in \Cref{tab_sim_topic_distribution}.
            Their similarity should be high since they have similar semantics.
            We see that TSCTM has the highest similarity among all models.
            These are because TSCTM can sufficiently model the relations among samples with the novel topic-semantic contrastive learning, which refines the representations and thus topic distributions.
            These results can further demonstrate the effectiveness of our proposed topic-semantic contrastive learning method.

    \section{Conclusion}
        In this paper, we propose TSCTM, a novel and unified method for topic modeling of short texts.
        Our method with the novel topic-semantic contrastive learning can refine the learning of representations through sufficiently modeling the relations among texts, regardless of the data augmentation availability.
        Experiments show our model effectively alleviates the data sparsity issue and consistently outperforms state-of-the-art baselines, generating high-quality topics and deriving useful topic distributions of short texts.

    \input{input/sec_limitations}

    \section*{Acknowledgement}
        We want to thank all anonymous reviewers for their helpful comments.
        This work was supported by Alibaba Innovative Research (AIR) programme with research grant AN-GC-2021-005.

    \bibliography{library}

    \input{input/sec_appendix}

\end{document}

%% file: input/fig_motivation.tex
\begin{figure}
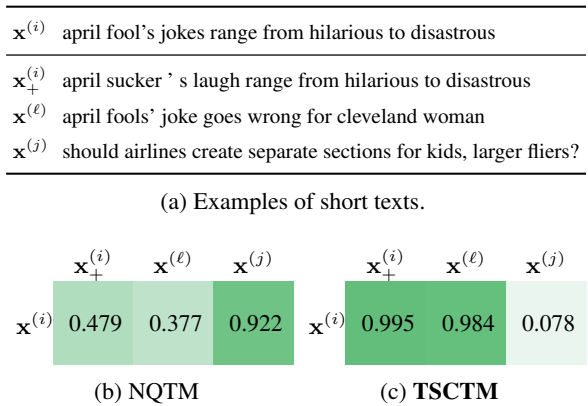

    \centering
    \begin{subfigure}[b]{\linewidth}
        \centering
        \small
        \setlength{\tabcolsep}{1mm}
        \renewcommand{\arraystretch}{1.5}
        \resizebox{\linewidth}{!}{
        \begin{tabular}{ll}
            \toprule
            $\mbf{x}^{(i)}$   & april fool's jokes range from hilarious to disastrous \\
            \midrule
            $\mbf{x}^{(i)}_{+}$  & april sucker ' s laugh range from hilarious to disastrous \\
            $\mbf{x}^{(\ell)}$   & april fools' joke goes wrong for cleveland woman \\
            $\mbf{x}^{(j)}$   & should airlines create separate sections for kids, larger fliers? \\
            \bottomrule
        \end{tabular}%
        }
        \caption{Examples of short texts.}
        \label{fig_motivation_texts}
    \end{subfigure}%
    \vspace{1em}
    \\
    \begin{subfigure}[b]{0.5\linewidth}
        \centering
        \small
        \begin{tabular}{p{0.85em} p{2em} p{2em} p{2em}}
                & \multicolumn{1}{c}{$\mbf{x}^{(i)}_{+}$}     & \multicolumn{1}{c}{$\mbf{x}^{(\ell)}$} & \multicolumn{1}{c}{$\mbf{x}^{(j)}$} \\
            \parbox[c][3.5em]{0.85em}{\centering $\mbf{x}^{(i)}$}   & \parbox[c][3.5em]{2em}{\centering  \cellcolor[rgb]{ .686,  .867,  .741}0.479} & \parbox[c][3.5em]{2em}{\centering \cellcolor[rgb]{ .745,  .89,  .792}0.377} & \parbox[c][3.5em]{2em}{\centering \cellcolor[rgb]{ .431,  .765,  .522}0.922} \\
        \end{tabular}%
        \caption{NQTM}
        \label{fig_motivation_baseline}
    \end{subfigure}%
    \begin{subfigure}[b]{0.5\linewidth}
        \centering
        \small
        \begin{tabular}{p{0.85em} p{2em} p{2em} p{2em}}
                & \multicolumn{1}{c}{$\mbf{x}^{(i)}_{+}$}     & \multicolumn{1}{c}{$\mbf{x}^{(\ell)}$} & \multicolumn{1}{c}{$\mbf{x}^{(j)}$} \\
                \parbox[c][3.5em]{0.85em}{\centering $\mbf{x}^{(i)}$}   & \parbox[c][3.5em]{2em}{\centering  \cellcolor[rgb]{ .388,  .745,  .482}0.995} & \parbox[c][3.5em]{2em}{\centering  \cellcolor[rgb]{ .396,  .749,  .49}0.984} & \parbox[c][3.5em]{2em}{\centering  \cellcolor[rgb]{ .918,  .961,  .937}0.078} \\
        \end{tabular}%
        \caption{\textbf{TSCTM}}
        \label{fig_motivation_our}
    \end{subfigure}%
    \caption{
        (a):
        Examples of short texts from TagMyNews title dataset.
        Text $\mbf{x}^{(i)}_{+}$ is an augmented view of $\mbf{x}^{(i)}$, and $\mbf{x}^{(i)}$ and $\mbf{x}^{(\ell)}$ are relevant while $\mbf{x}^{(i)}$ and $\mbf{x}^{(j)}$ are irrelevant.
        (b, c):
        Heat map of cosine similarity between learned topic distributions.
        The similarities of our TSCTM are more reasonable than NQTM.
    }
    \vspace{-1em}
    \label{fig_motivation}
\end{figure}

%% file: input/sec_related_work.tex
\section{Related Work} \label{sec_related_work}

    \paragraph{Topic Modeling}
        Based on classic long text topic models \cite{hofmann1999probabilistic,blei2003latent,lee2020prior},
        various probabilistic topic models for short texts have been proposed \cite{yan2013biterm,yin2014dirichlet,Li2016,Wu2019}.
        They use Gibbs Sampling \cite{griffiths2004finding} or Variational Inference \cite{blei2017variational} to infer model parameters.
        Later, due to the effectiveness and brevity of Variational AutoEncoder \cite[VAE, ][]{Kingma2014a,Rezende2014}, many neural topic models have been introduced \cite{Miao2016,Miao2017,Srivastava2017,Card2018a,Nan2019,dieng2020topic,li2021topic,Wu2020,Wu2020short,wu2021discovering,wang2021extracting}.
        Among those methods, the most related one to this paper is NQTM \cite{Wu2020short}.
        Although NQTM also uses vector quantization to aggregate the short texts with similar topics,
        however, we note that our method differs significantly in that:
        \begin{inparaenum}[(i)]
            \item
                Our TSCTM framework uses the novel topic-semantic contrastive learning method that fully considers the relations among samples with effective positive and negative sampling strategies, while NQTM only considers the relations between samples with similar semantics.
            \item
                Our TSCTM framework can adapt to the case with data augmentation by sufficiently modeling the relations brought by augmented samples, achieving higher performance gains, while NQTM cannot fully incorporate such relations.
        \end{inparaenum}

    \paragraph{Contrastive Learning}
        The idea of contrastive learning is to measure the similarity relations of sample pairs in a representation space \cite{hadsell2006dimensionality,Song2016deep,hjelm2018learning,van2018representation,frosst2019analyzing,wang2019multi,he2020momentum,wang2020understanding}.
        It has been widely explored in the visual field, such as
        image classification \cite{chen2020simple,khosla2020supervised}, 
        objective detection \cite{xie2021detco}, and image segmentation \cite{zhao2021contrastive}.
        For text data,
        some studies use contrastive loss \cite{gao2021simcse,nguyen2021contrastive} by sampling salient words from texts to build positive samples, but they could be inappropriate for short text topic modeling due to the limited context of short texts (shown in \Cref{sec_topic_quality}).
        In contrast, our new framework can
        discover effective samples
        for learning contrastively based on the topic semantics  and can smoothly adapt to the case with augmentations,
        both of which better fit the short text modeling context.

%% file: input/fig_illustration.tex
\begin{figure}[!t]
    \centering
    \includegraphics[width=\linewidth]{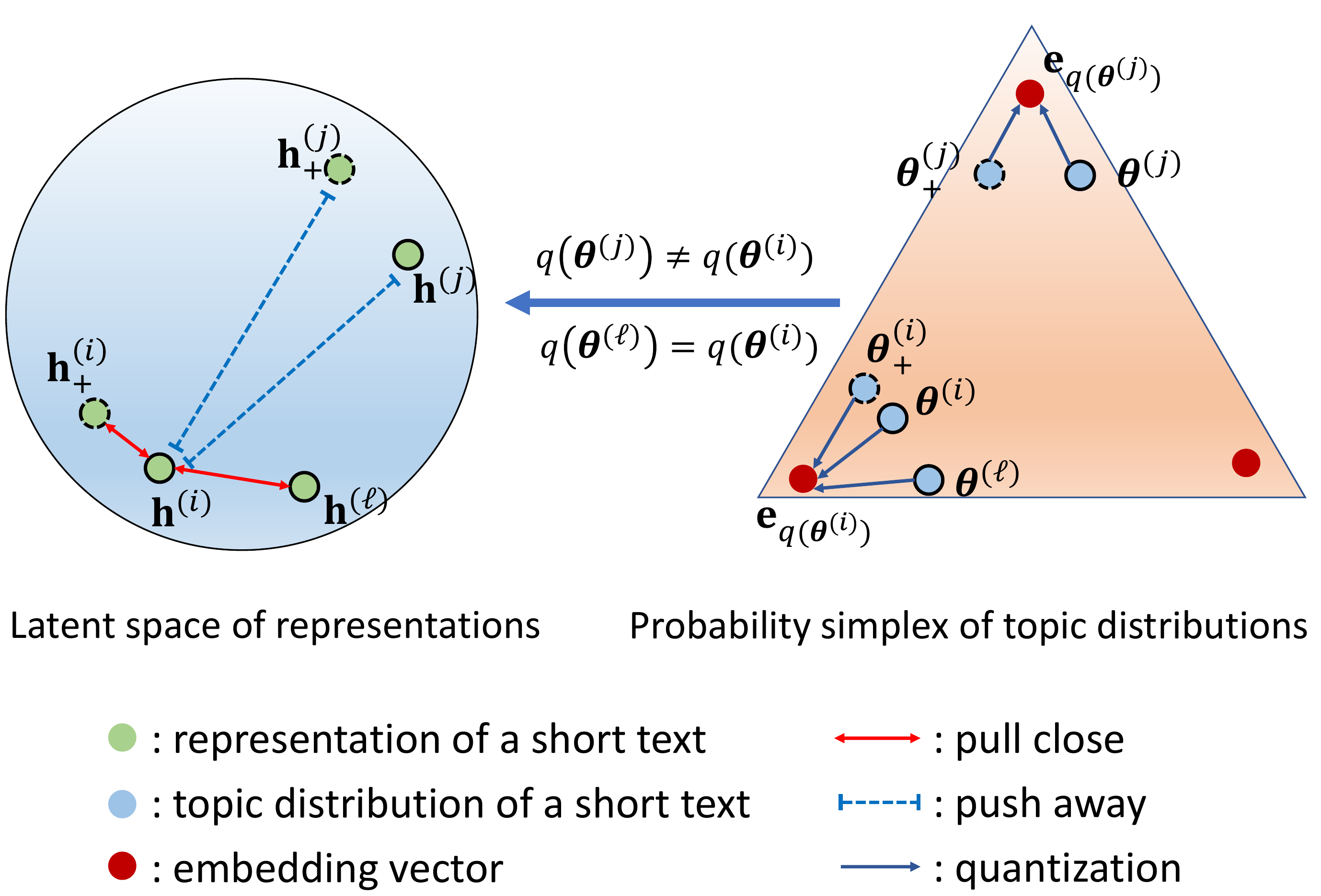}
    \caption{
        Illustration of the proposed topic-semantic contrastive learning.
        It refines the learning of representations through modeling the relations of samples according to their topic semantics
        (only solid line circles exist when without data augmentation).
    }
    \label{fig_illustration}
\end{figure}

%% file: input/tab_dataset.tex
\begin{table}[!t]
    \centering
    \small
        \begin{tabular}{lrrr}
        \toprule
        \multicolumn{1}{c}{\multirow{2}[2]{*}{Dataset}} & \# of & Vocabulary & \#labels \\
              & texts & Size  &  \\
        \midrule
        TagMyNews title & 31,223  & 6,391  & 7 \\
        AG News & 8,000  & 5,603  & 4 \\
        Google News & 11,066  & 2,451  & 152 \\
        \bottomrule
        \end{tabular}%
    \caption{Dataset statistics.}
    \label{tab_dataset}
\end{table}

%% file: input/tab_topic_quality.tex
\begin{table*}[!ht]
    \centering
    \setlength{\tabcolsep}{2mm}
    \renewcommand{\arraystretch}{1.2}
    \small
    \begin{tabular}{lrrrrrrrrrrrrrr}
    \toprule
    \multicolumn{1}{c}{\multirow{3}[4]{*}{Model}} & \multicolumn{4}{c}{TagMyNews title} &       & \multicolumn{4}{c}{AG News}   &       & \multicolumn{4}{c}{Google News} \\
    \cmidrule{2-5}\cmidrule{7-10}\cmidrule{12-15}      & \multicolumn{2}{c}{$K$=50} & \multicolumn{2}{c}{$K$=100} &       & \multicolumn{2}{c}{$K$=50} & \multicolumn{2}{c}{$K$=100} &       & \multicolumn{2}{c}{$K$=50} & \multicolumn{2}{c}{$K$=100} \\
          & \multicolumn{1}{c}{$C_V$} & \multicolumn{1}{c}{$TU$} & \multicolumn{1}{c}{$C_V$} & \multicolumn{1}{c}{$TU$} &       & \multicolumn{1}{c}{$C_V$} & \multicolumn{1}{c}{$TU$} & \multicolumn{1}{c}{$C_V$} & \multicolumn{1}{c}{$TU$} &       & \multicolumn{1}{c}{$C_V$} & \multicolumn{1}{c}{$TU$} & \multicolumn{1}{c}{$C_V$} & \multicolumn{1}{c}{$TU$} \\
    \midrule
    \multicolumn{15}{l}{\textbf{Without Data Augmentation}} \\
    ProdLDA & 0.397 & 0.929 & 0.420 & 0.894 &       & 0.451 & 0.563 & 0.449 & 0.610 &       & 0.417 & 0.725 & 0.405 & 0.655 \\
    WLDA  & 0.361 & 0.740 & 0.360 & 0.634 &       & 0.387 & 0.585 & 0.384 & 0.507 &       & 0.376 & 0.736 & 0.366 & 0.604 \\
    CLNTM & 0.352 & 0.556 & 0.320 & 0.246 &       & 0.439 & 0.722 & 0.426 & 0.620 &       & 0.416 & 0.699 & 0.409 & 0.641 \\
    NQTM  & 0.432 & 0.985 & 0.424 & 0.932 &       & 0.408 & 0.977 & 0.406 & 0.920 &       & 0.405 & 0.951 & 0.390 & 0.889 \\
    WeTe  & 0.376 & 0.865 & 0.304 & 0.609 &       & 0.410 & 0.966 & 0.380 & 0.853 &       & 0.388 & 0.922 & 0.331 & 0.681 \\
    \midrule
    \textbf{TSCTM} & \textbf{0.445} & \textbf{0.997} & \textbf{0.456} & \textbf{0.936} &       & \textbf{0.460} & \textbf{0.990} & \textbf{0.452} & \textbf{0.954} &       & \textbf{0.424} & \textbf{0.995} & \textbf{0.426} & \textbf{0.934} \\
    \midrule
    \midrule
    \multicolumn{15}{l}{\textbf{With Data Augmentation}} \\
    ProdLDA & 0.411 & 0.950 & 0.433 & 0.920 &       & 0.472 & 0.623 & 0.473 & 0.528 &       & 0.416 & 0.777 & 0.399 & 0.742 \\
    WLDA  & 0.354 & 0.779 & 0.354 & 0.692 &       & 0.378 & 0.811 & 0.373 & 0.697 &       & 0.356 & 0.753 & 0.352 & 0.603 \\
    CLNTM & 0.309 & 0.244 & 0.309 & 0.115 &       & 0.462 & 0.647 & 0.448 & 0.451 &       & 0.453 & 0.412 & 0.412 & 0.385 \\
    NQTM  & 0.458 & 0.995 & 0.464 & 0.930 &       & 0.422 & 0.982 & 0.420 & 0.939 &       & 0.404 & 0.964 & 0.382 & 0.902 \\
    WeTe  & 0.372 & 0.905 & 0.331 & 0.733 &       & 0.410 & 0.991 & 0.374 & 0.818 &       & 0.365 & 0.862 & 0.319 & 0.647 \\
    \midrule
    \textbf{TSCTM} & \textbf{0.514} & \textbf{0.997} & \textbf{0.509} & \textbf{0.968} &       & \textbf{0.493} & \textbf{0.996} & \textbf{0.479} & \textbf{0.969} &       & \textbf{0.467} & \textbf{1.000} & \textbf{0.446} & \textbf{0.968} \\
    \bottomrule
    \end{tabular}%
    \caption{
      Topic coherence ($C_V$) and diversity ($TU$) results under 50 and 100 topics ($K \!\! = \!\! 50$ and $K \!\!=\!\! 100$).
      \textbf{Without Data Augmentation} means only original datasets are used, and \textbf{With Data Augmentation} means the augmented texts are used to enrich datasets for each model, so all models are evaluated in the same data conditions under two scenarios.
      The best scores are in \textbf{bold}.
    }
    \label{tab_topic_quality}%
\end{table*}%

%% file: input/tab_ablation_study.tex
\begin{table*}[!t]
    \centering
    \small
    \setlength{\tabcolsep}{1.5mm}
    \renewcommand{\arraystretch}{1.2}
    \resizebox{0.9\linewidth}{!}{
        \begin{tabular}{lrrrrrrrrrrrrrr}
        \toprule
        \multicolumn{1}{c}{\multirow{3}[4]{*}{Model}} & \multicolumn{4}{c}{TagMyNews title} &       & \multicolumn{4}{c}{AG News}   &       & \multicolumn{4}{c}{Google News} \\
        \cmidrule{2-15}      & \multicolumn{2}{c}{$K$=50} & \multicolumn{2}{c}{$K$=100} &       & \multicolumn{2}{c}{$K$=50} & \multicolumn{2}{c}{$K$=100} &       & \multicolumn{2}{c}{$K$=50} & \multicolumn{2}{c}{$K$=100} \\
              & \multicolumn{1}{c}{$C_V$} & \multicolumn{1}{c}{$TU$} & \multicolumn{1}{c}{$C_V$} & \multicolumn{1}{c}{$TU$} &       & \multicolumn{1}{c}{$C_V$} & \multicolumn{1}{c}{$TU$} & \multicolumn{1}{c}{$C_V$} & \multicolumn{1}{c}{$TU$} &       & \multicolumn{1}{c}{$C_V$} & \multicolumn{1}{c}{$TU$} & \multicolumn{1}{c}{$C_V$} & \multicolumn{1}{c}{$TU$} \\
        \midrule
        w/ traditional contrastive & 0.441 & 0.849 & 0.423 & 0.693 &       & 0.483 & 0.957 & 0.448 & 0.816 &       & 0.452 & 0.943 & 0.439 & 0.689 \\
        w/o negative pairs & 0.409 & 0.929 & 0.400 & 0.850 &       & 0.422 & 0.758 & 0.397 & 0.503 &       & 0.436 & 0.757 & 0.414 & 0.654 \\
        w/o positive pairs & 0.479 & 0.993 & 0.477 & 0.931 &       & 0.472 & 0.991 & 0.454 & 0.960 &       & 0.459 & 0.999 & 0.438 & 0.956 \\
        \midrule
        \textbf{TSCTM} & \textbf{0.514} & \textbf{0.997} & \textbf{0.509} & \textbf{0.968} &       & \textbf{0.493} & \textbf{0.996} & \textbf{0.479} & \textbf{0.969} &       & \textbf{0.467} & \textbf{1.000} & \textbf{0.446} & \textbf{0.968} \\
        \bottomrule
        \end{tabular}%

    }
    \caption{
        Ablation study of removing positive and negative pairs in the TSCTM (w/o negative pairs and w/o positive pairs), and using the traditional contrastive loss (w/ traditional contrastive).
        The best scores are in \textbf{bold}.
    }
    \label{tab_ablation_study}
\end{table*}

%% file: input/tab_clustering.tex
\begin{table}[!t]
    \centering
    \setlength{\tabcolsep}{1.2mm}
    \renewcommand{\arraystretch}{1.35}
    \resizebox{\linewidth}{!}{
        \begin{tabular}{lcccccccc}
            \toprule
            \multicolumn{1}{c}{\multirow{2}[4]{*}{Model}} & \multicolumn{2}{c}{TagMyNews title} &       & \multicolumn{2}{c}{AG News} &       & \multicolumn{2}{c}{Google News} \\
            \cmidrule{2-3}\cmidrule{5-6}\cmidrule{8-9}      & Purity & NMI   &       & Purity & NMI   &       & Purity & NMI \\
            \midrule
            ProdLDA & 0.260 & 0.002 &       & 0.773 & 0.267 &       & 0.089 & 0.137 \\
            WLDA  & 0.363 & 0.058 &       & 0.583 & 0.148 &       & 0.411 & 0.608 \\
            CLNTM & 0.266 & 0.008 &       & 0.408 & 0.097 &       & 0.099 & 0.136 \\
            NQTM  & 0.595 & 0.231 &       & 0.800 & 0.310 &       & 0.555 & 0.753 \\
            WeTe  & 0.487 & 0.180 &       & 0.713 & 0.307 &       & 0.301 & 0.560 \\
            \midrule
            \textbf{TSCTM} & \textbf{0.610} & \textbf{0.239} &       & \textbf{0.811} & \textbf{0.317} &       & \textbf{0.563} & \textbf{0.766} \\
            \bottomrule
        \end{tabular}%
    }
    \caption{Text clustering results of Purity and NMI. The best scores of each dataset are highlighted in \textbf{bold}.}
    \label{tab_clustering}%
\end{table}%

%% file: input/fig_classification.tex
\begin{figure}[!t]
    \centering
    \includegraphics[width=\linewidth]{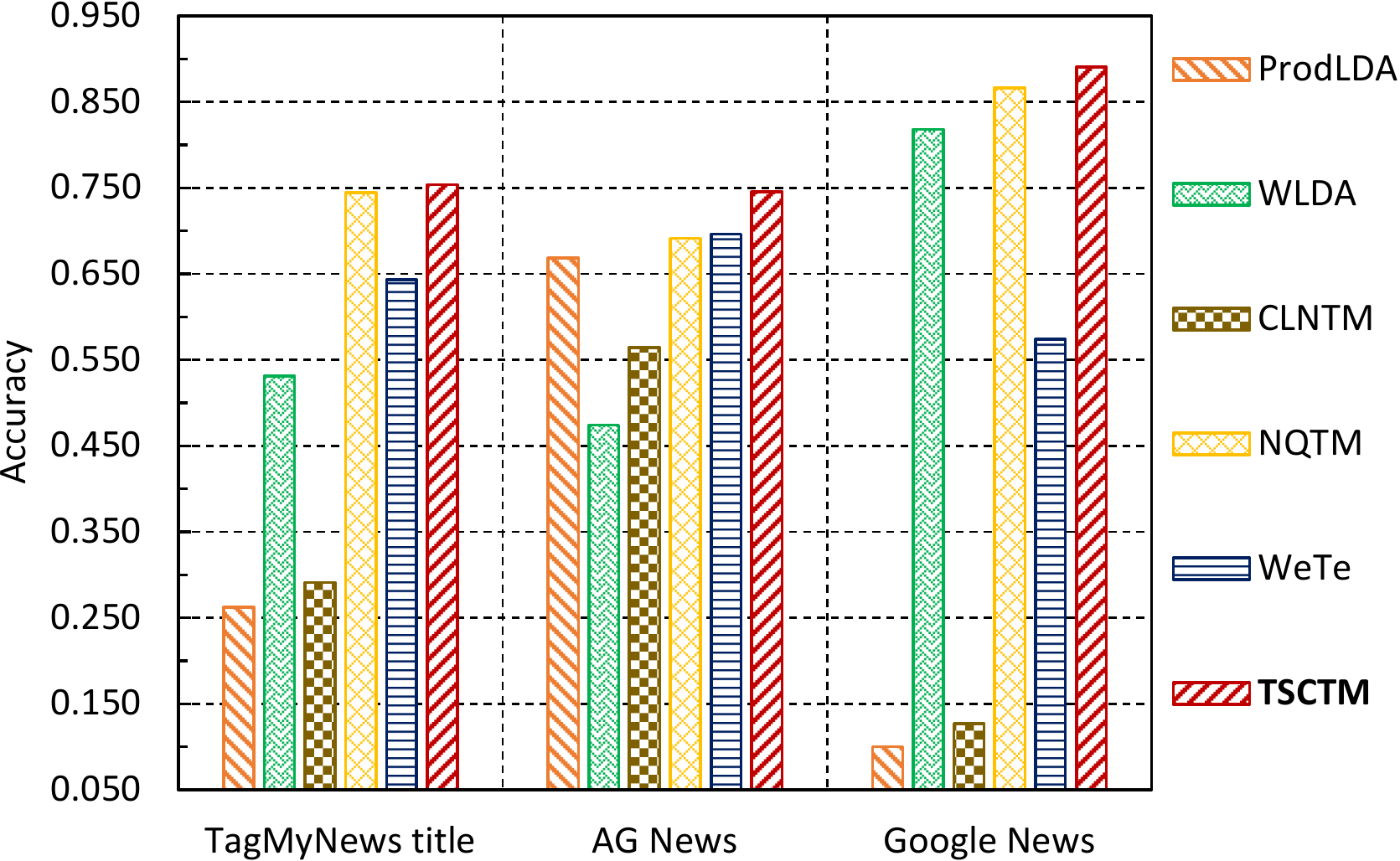}
    \caption{Text classification results with topic distributions learned by topic models.}
    \label{fig_classification}
\end{figure}

%% file: input/fig_tsne.tex
\begin{figure*}
    \centering
    \begin{subfigure}[b]{0.333\linewidth}
        \centering
        \includegraphics[width=0.8\linewidth]{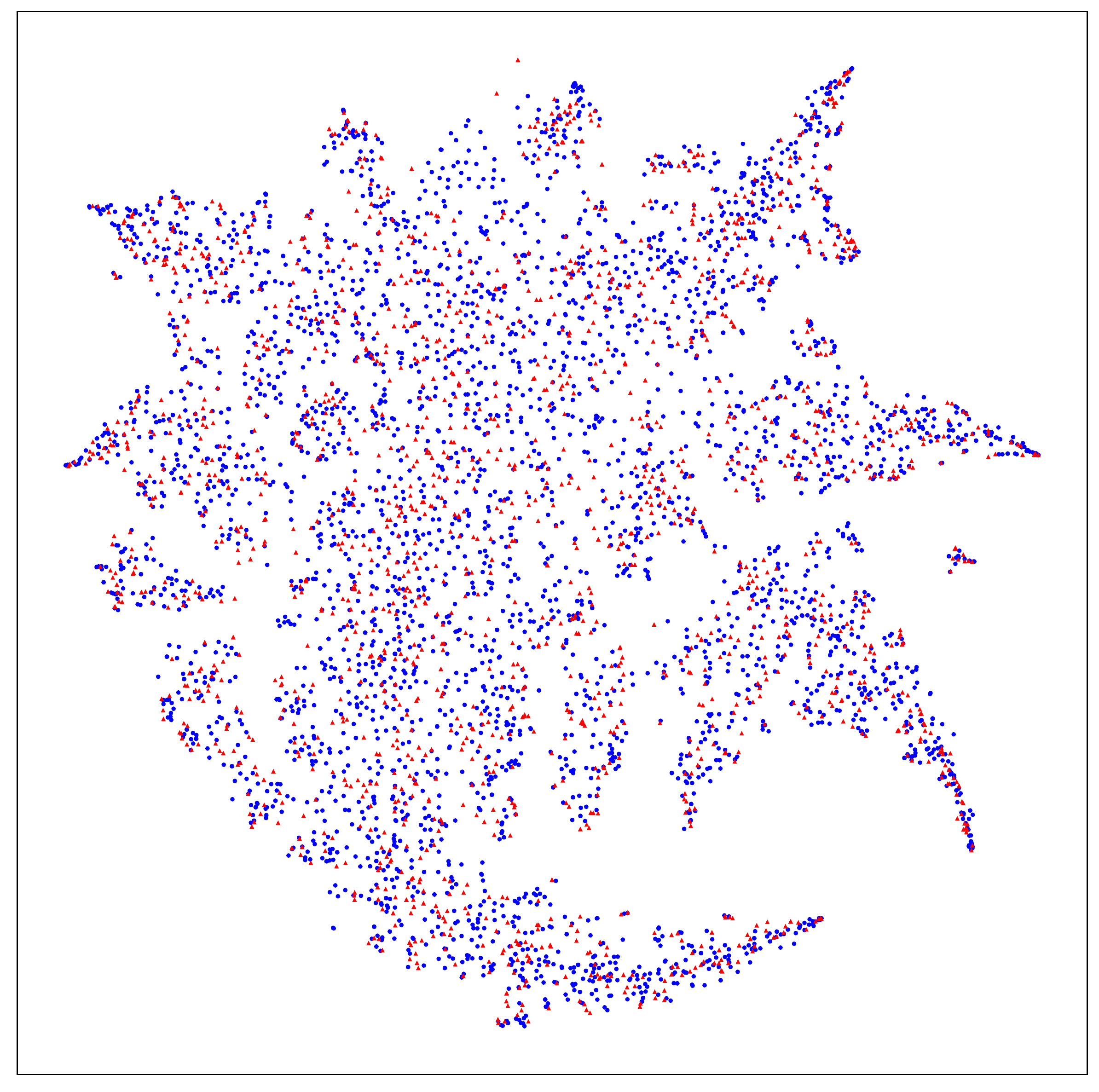}
        \caption{ProdLDA}
        \label{fig_tsne_ProdLDA}
    \end{subfigure}%
    \begin{subfigure}[b]{0.333\linewidth}
        \centering
        \includegraphics[width=0.8\linewidth]{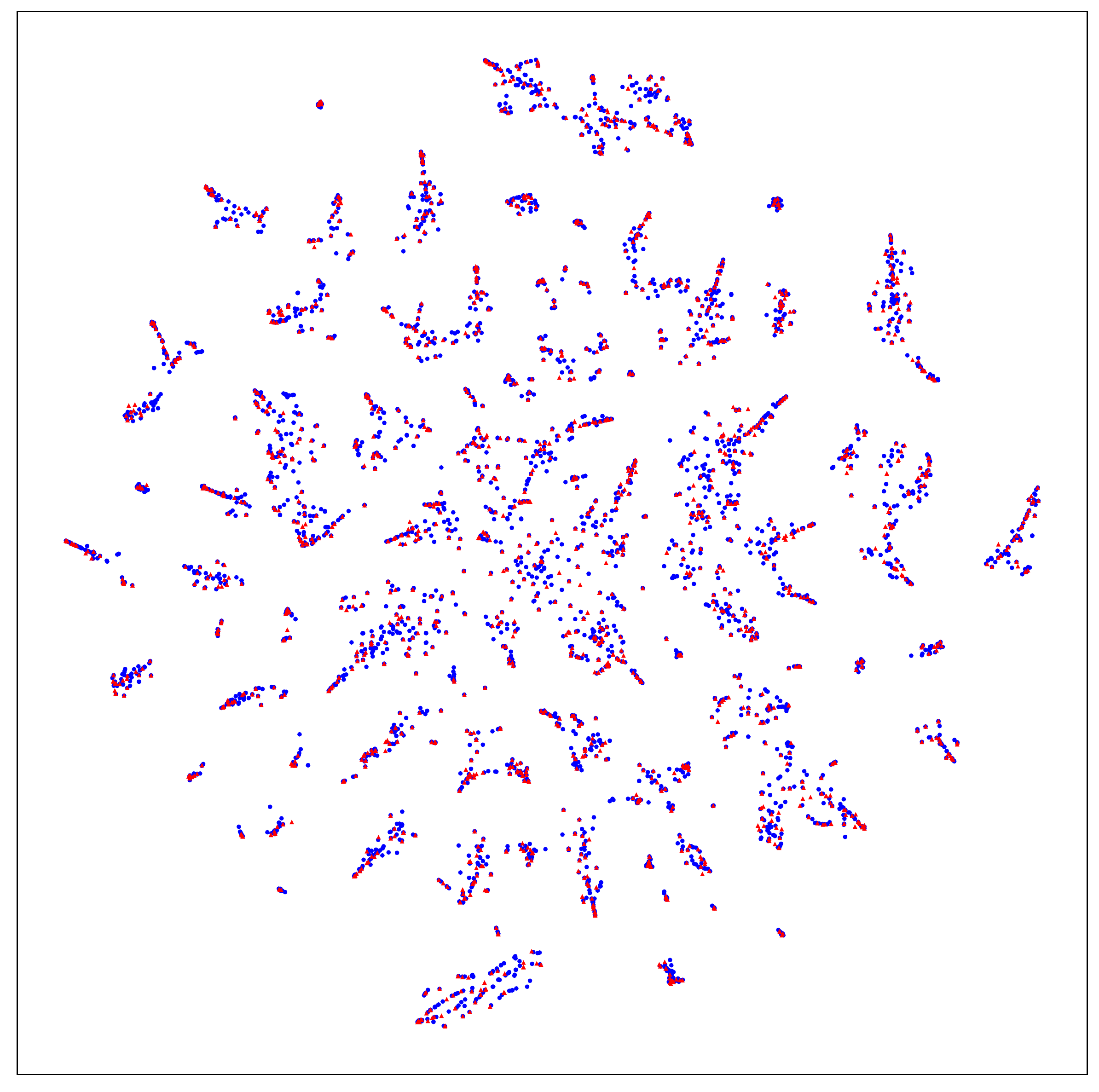}
        \caption{NQTM}
        \label{fig_tsne_NQTM}
    \end{subfigure}%
    \begin{subfigure}[b]{0.333\linewidth}
        \centering
        \includegraphics[width=0.8\linewidth]{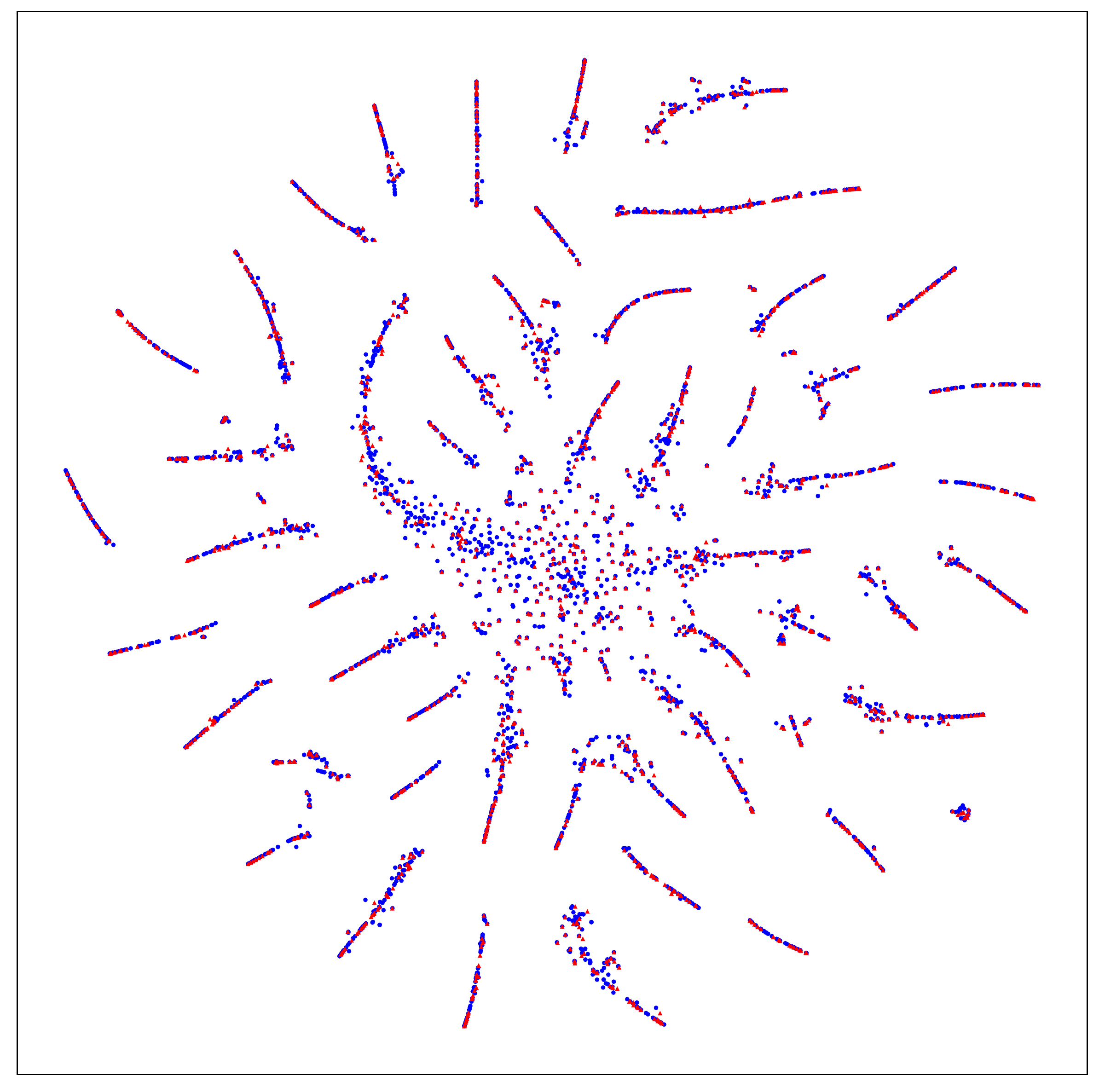}
        \caption{\textbf{TSCTM}}
        \label{fig_tsne_TSCTM}
    \end{subfigure}%
    \caption{
        t-SNE visualization of learned topic distributions of original ({\color{red} $\blacktriangle$}) and augmented ({\color{blue} $\bullet$}) short texts.
        Compared to ProdLDA and NQTM,
        the points of TSCTM are better aggregated and separately scattered in the space.
    }
    \label{fig_tsne}
\end{figure*}

%% file: input/tab_sim.tex
\begin{table}[!t]
    \centering
    \small
    \setlength{\tabcolsep}{1.2mm}
    \renewcommand{\arraystretch}{1.35}
    \resizebox{\linewidth}{!}{
        \begin{tabular}{lrrrrrrrr}
        \toprule
        \multicolumn{1}{c}{\multirow{2}[4]{*}{Model}} & \multicolumn{2}{c}{TagMyNews title} &       & \multicolumn{2}{c}{AG News} &       & \multicolumn{2}{c}{Google News} \\
        \cmidrule{2-3}\cmidrule{5-6}\cmidrule{8-9}      & \multicolumn{1}{c}{$K$=50} & \multicolumn{1}{c}{$K$=100} &       & \multicolumn{1}{c}{$K$=50} & \multicolumn{1}{c}{$K$=100} &       & \multicolumn{1}{c}{$K$=50} & \multicolumn{1}{c}{$K$=100} \\
        \midrule
        ProdLDA & 0.412 & 0.272 &       & 0.612 & 0.549 &       & 0.304 & 0.433 \\
        WLDA  & 0.640 & 0.620 &       & 0.860 & 0.866 &       & 0.887 & 0.870 \\
        CLNTM & 0.541 & 0.477 &       & 0.425 & 0.405 &       & 0.572 & 0.511 \\
        NQTM  & 0.870 & 0.858 &       & 0.962 & 0.963 &       & 0.947 & 0.942 \\
        WeTe  & 0.839 & 0.816 &       & 0.938 & 0.926 &       & 0.914 & 0.907 \\
        \midrule
        \textbf{TSCTM} & \textbf{0.946} & \textbf{0.945} &       & \textbf{0.986} & \textbf{0.987} &       & \textbf{0.974} & \textbf{0.974} \\
        \bottomrule
        \end{tabular}%
    }
    \caption{
        Cosine similarity between topic distributions of original and augmented short texts.
        The highest are in \textbf{bold}.
    }
    \label{tab_sim_topic_distribution}
\end{table}

%% file: input/sec_limitations.tex
\section*{Limitations}
Our method achieves promising performance to mitigate data sparsity for short text topic modeling,
but we believe that there are two limitations to be explored for future works:
\begin{inparaenum}[(i)]
    \item
        More data augmentation techniques may be studied to further improve short text topic modeling performance.
    \item
        The possible metadata of short texts, like authors, hashtags, and sentiments, can be considered to further assist the modeling of relations.
\end{inparaenum}

%% file: input/sec_appendix.tex
\clearpage

    \input{input/tab_topic_examples}

\appendix

\section{Model Implementation} \label{sec_appendix_model_implementation}
    We conduct experiments on NVIDIA GPU, and it takes less than 0.5 GPU hours to train our model on each dataset.
    For our model, the encoder network $f_{\Theta}$ is a two-layer MLP with softplus as the activation function, same as \citet{Wu2020short}, and we use Adam \cite{Kingma2014} to optimize model parameters.
    We run our model for 200 epochs with
    learning rate as 0.002 following \citet{Srivastava2017}, and $\lambda$ as 0.1 following \citet{VandenOord2017}.

\section{Examples of Discovered Topics} \label{sec_appendix_topic_examples}
    Following \citet{Nan2019,Wu2020short}, we randomly select some examples of discovered topics by ProdLDA, NQTM, and our TSCTM from Google News for qualitative study since they perform relatively better among baselines.
    As shown in \Cref{tab_topic_examples},
    ProdLDA produces several redundant topics including ``giraffe'', and these topics are less informative as they are associated with irrelevant words like ``fundamentalist'' and ``animation''.
    NQTM also has repetitive topics about ``kanye''.
    In contrast, our TSCTM only generates one coherent topic about ``animation'', ``kanye'', and ``giraffe'' with relevant words.
    For example, the topic of TSCTM is more focused on ``animation'' with ``disney'', the movie name ``frozen'' and its theme song singer ``idina menzel''.

%% file: input/tab_topic_examples.tex
\begin{table*}[!ht]
    \centering
    \small
    \renewcommand{\arraystretch}{1.3}
        \begin{tabular}{ll}
            \toprule
            Models & Examples of Topics \\
            \midrule
            \multirow{3}[1]{*}{ProdLDA}
                & perrish apps chart \uline{giraffe} cleared lash mary tyrese fill fundamentalist \\ %
                & blog mistake duel reduce sleet \uline{giraffe} animation tradition stress freezing \\ %
                & major \uline{giraffe} offence moment halo lifetime jim sharing draft congo \\ %
            \midrule
            \multirow{3}[2]{*}{NQTM}
                & \uline{kanye} \uline{west} confirms yeezus adidas leaf album rant concert kravitz \\ %
                & kim \uline{west} \uline{kanye} invited brody jenner kardashian wedding beautiful invite \\ %
                & \uline{kanye} james video \uline{west} bound recreate kimye franco shot music \\ %
            \midrule
            \multirow{3}[2]{*}{\textbf{TSCTM}}
                & giraffe congo poaching forgotten habitat ape okapi bonobo specie endangered \\ %
                & frozen disney animation idina menzel kristen animated melt fairy bell \\ %
                & adidas nike partnership summer lenny kravitz confirms cruel kanye album \\ %
            \bottomrule
        \end{tabular}%
    \caption{
        Top 10 related words of discovered topics from Google News.
        Repetitive words are \uline{underlined}.
    }
    \label{tab_topic_examples}
\end{table*}